\pdfoutput=1
\documentclass[10pt, a4paper]{article}

\usepackage{lrec-coling2024} 
\usepackage{amsmath}
\usepackage{tabularx}
\usepackage{xcolor}

\definecolor{darkgreen}{RGB}{0, 150, 0} 
\definecolor{darkred}{RGB}{150, 0, 0} 

\usepackage{natbib}
\usepackage{multibib}
\makeatletter
\def\@mb@citenamelist{cite,citep,citet,citealp,citealt,citepalias,citetalias}
\makeatother
\newcites{languageresource}{~}

\usepackage{graphicx}
\usepackage{tabularx}
\usepackage{soul}
\usepackage{titlesec}
\titleformat{\section}{\normalfont\large\bfseries\center}{\thesection.}{1em}{}
\titleformat{\subsection}{\normalfont\SmallTitleFont\bfseries\raggedright}{\thesubsection.}{1em}{}
\titleformat{\subsubsection}{\normalfont\normalsize\bfseries\raggedright}{\thesubsubsection.}{1em}{}
\renewcommand\thesection{\arabic{section}}
\renewcommand\thesubsection{\thesection.\arabic{subsection}}
\renewcommand\thesubsubsection{\thesubsection.\arabic{subsubsection}}

\usepackage{xcolor}
\usepackage{hyperref}
 \definecolor{darkblue}{rgb}{0, 0, 0.5}
  \hypersetup{colorlinks=true, citecolor=darkblue, linkcolor=darkblue, urlcolor=darkblue}

\usepackage{xstring}

\usepackage{color}

\title{Fisher Mask Nodes for Language Model Merging}

\name{Thennal D K, Ganesh Nathan, Suchithra M S} 

\address{
\\ Indian Institute of Information Technology Kottayam 
\\ Kerala, India
\\ \{thennal21bcs14, ganesh21bcs104, suchithra\}@iiitkottayam.ac.in}

\abstract{
Fine-tuning pre-trained models provides significant advantages in downstream performance. The ubiquitous nature of pre-trained models such as BERT and its derivatives in natural language processing has also led to a proliferation of task-specific fine-tuned models. As these models typically only perform one task well, additional training or ensembling is required in multi-task scenarios. The growing field of model merging provides a solution, dealing with the challenge of combining multiple task-specific models into a single multi-task model. 
In this study, we introduce a novel model merging method for Transformers, combining insights from previous work in Fisher-weighted averaging and the use of Fisher information in model pruning. Utilizing the Fisher information of mask nodes within the Transformer architecture, we devise a computationally efficient weighted-averaging scheme. Our method exhibits a regular and significant performance increase across various models in the BERT family, outperforming full-scale Fisher-weighted averaging in a fraction of the computational cost, with baseline performance improvements of up to +6.5 and a speedup between 57.4x and 321.7x across models. Our results prove the potential of our method in current multi-task learning environments and suggest its scalability and adaptability to new model architectures and learning scenarios.
 \\ \newline  }

\begin{document}

\maketitleabstract

\section{Introduction}

In recent years, pre-trained models have become ubiquitous in natural language processing (NLP) \citep{pretrained, pretrained2, zhou2023comprehensive, min2021recent}. In particular, models inheriting the Transformer architecture \citep{vaswani2017attention} such as BERT and its variants \citep{devlin2019bert} are popular for a wide variety of tasks \citep{zhou2023comprehensive, min2021recent}. By finetuning on downstream tasks, pre-trained models achieve improved performance on those specific tasks. However, individual fine-tuned models for each task necessitates significant overhead in multi-task scenarios, as multiple separate models need to be running at once \citep{zhang2021survey, fifty2021efficiently, ties}. Furthermore, individual fine-tuned models sacrifice generalization capabilities \citep{jin2023dataless, arpit2022ensemble, ties, taskvectors}. Fine-tuning on multiple tasks at once is a possible solution, but is generally resources-intensive and costly \citep{ties, fifty2021efficiently}.

 Model merging is a promising field that can alleviate this problem. Model merging predominantly deals with the challenges of fusing multiple separate models—generally of the same architecture but with different parameters—into a single model. A growing body of literature addresses the restricted context of merging multiple task-specific models into a single multitask model \citep{ties, matena2022merging, taskvectors, jin2023dataless}. Fisher-weighted averaging \citep{matena2022merging} is one such method that uses the diagonal approximation of the Fisher information of each parameter as weights in a weighted-averaging scheme. However, doing so requires access to the validation set that the model will be evaluated on, along with significant resource overhead, as gradients for all parameters are required to calculate the approximated Fisher information.

Fisher information has been used extensively in machine learning as a measure of parameter importance \citep{pruning, liu2021grouppruning, catastrophic, pascanu2014revisiting, soen2021variance}. In recent years, it has gained particular prominence in the field of model pruning \citep{liu2021grouppruning, pruning}. \citet{pruning} devised a pruning scheme for Transformer models utilizing Fisher information matrices. Masks are inserted in every feed-forward layer and attention head, and the approximated Fisher information of the masks is used as an indication of the necessity of the affected block of parameters. Low-importance parameters, indicated by this metric, are then pruned.

Our work focuses on modifying the Fisher-weighted merging method devised by \citet{matena2022merging} with the insights gained from \citet{pruning}: that the Fisher information of masks can act as an importance measure for their affected parameter blocks. We insert masks and calculate an approximate of the Fisher information of each mask. By associating the mask Fisher information to relevant parameters in an empirical scheme, we merge models via weighted averaging similar to \citet{matena2022merging}. By only calculating the Fisher information of masks instead of the Fisher information of all parameters, our methods provide a massive reduction in computational cost while improving the performance benefits. As we restrict our context to performance on the respective tasks of fine-tuned model merges, our method does not require access to the validation set. The Fisher information matrix of each fine-tuned model on its respective task can provide a measure of how important each parameter is to that particular model, and we hypothesize that weighing by said Fisher information allows constructive merging, resulting in a single model performant in all given tasks.

We validate our hypothesis by demonstrating the effectiveness of our proposed methodology on various architectures of differing sizes, combining finetuned models on six tasks from the GLUE benchmark \citep{wang-etal-2018-glue}. Our method provides a significant improvement over simple averaging and Fisher-weighted merging in most models, using extremely small amounts of data and computational resources.  We also note our method's efficiency at scale, with a speedup ranging from \textbf{57.4x} to \textbf{321.7x} across models in comparison to Fisher-weighted merging.

Our code is publicly released on Github.\footnote{\href{https://github.com/thennal10/fisher-nodes-merging}{https://github.com/thennal10/fisher-nodes-merging}}

\section{Related Work}

Much of the work in model merging has been done in addressing permutation symmetry when models are trained from different initializations \citep{ainsworth2023git, singh2023model, jin2023dataless, li2016convergent, tatro2020optimizing, entezari2022role}. However, in the case of different fine-tuned models trained from the same parent pre-trained model, merging can often be done directly. As the optimization trajectory overlaps at the beginning, even simple averaging can provide reasonable results \citep{choshen2022fusing, wortsman2022modelsoup, taskvectors, jin2023dataless}. In this setting, model merging has been explored for a wide variety of contexts including improving single-task performance \citep{choshen2022fusing, wortsman2022modelsoup}, transfer learning \citep{matena2022merging}, federated learning \citep{li2020convergence, McMahan2016CommunicationEfficientLO}, and improving generalization \citep{jin2023dataless, taskvectors, arpit2022ensemble, cha2021swad}. 

Restricting ourselves to the specific challenge of merging fine-tuned task-specific models for multi-task learning, many methods improve over the baseline of simple parameter averaging \citep{jin2023dataless, ties, taskvectors, matena2022merging}. Fisher-weighted averaging, proposed by \citet{matena2022merging}, was originally evaluated in the context of transfer learning but it has been further employed for multi-task learning \citep{ties}. 

The Fisher information matrix as a statistical measure has been used extensively in machine learning for a wide variety of applications \citep{catastrophic, pruning, pascanu2014revisiting, soen2021variance, hannun2021measuring, liu2021grouppruning}. \citet{liu2021grouppruning} prominently uses it for the task of pruning across multiple model architectures. \citet{pruning} provides a fast Transformer-specific pruning scheme using the Fisher information matrix of mask nodes.

As the full Fisher information matrix is infeasible to calculate for most applications, only the diagonals are computed over a specified sample as an approximation \citep{matena2022merging, catastrophic}. However, even the diagonal approximation incurs the same computational cost as training on a specified number of examples. Our method borrows the mask architecture from \citet{pruning}, applying it to the field of model merging to reduce computational cost while keeping the performance benefits of Fisher-weighted averaging. We also note that our method does not require access to the validation set, in contrast to other methods \citep{matena2022merging,ties,taskvectors}.

\section{Methodology}

Our approach relies on the hypothesis that the Fisher information of mask nodes can represent the importance of the parameters enclosed by the mask, as inspired by \citet{pruning}. Using the Fisher information of the mask nodes as a proxy for the Fisher information of the corresponding block of parameters, we merge parameters in a weighted-averaging merging scheme similar to \citet{matena2022merging}. We further consider only the training sets of the fine-tuned models for calculating the Fisher information in order to remove the dependency on the validation set and widen the applicability of our method.

In accordance with \citet{pruning}, our method focuses on the Transformer architecture, specifically the BERT architecture family \citep{devlin2019bert}. BERT can be characterized as a stack of homogeneous Transformer encoder blocks, consisting of a multi-headed attention block followed by a feed-forward block. Masks are inserted on each head of the multi-headed attention block, and on each row in the intermediate linear layer of the feed-forward block, referred to as filters by \citet{pruning}.

Following the notation of the formal BERT algorithm given by \citet{phuong2022formal}, we represent the modified multi-headed attention algorithm in Equation \ref{mha} and the modified feed-forward layer in Equation \ref{ffd}:

\begin{equation} \label{mha}
\begin{aligned}
&\text{For } h \in \{1,...,H\}: \\
& \quad Y_h = \texttt{Attention}(X; W^{h,l}_{qkv}) \\
& Y = \{Y_1, Y_2, ..., Y_H\} \\
& X \leftarrow X + W^o(Y \odot m^l_{mha}) + b^o \\
\end{aligned}
\end{equation}

\begin{equation} \label{ffd}
\begin{split}
X \leftarrow X + W^{l}_{mlp2}\texttt{GELU}(m^{l}_{mlp} \odot W^{l}_{mlp1} X + b^{l}_{mlp1}) \\ + b^{l}_{mlp2} 
\end{split}
\end{equation}
  
We assume a given input $X$ and present only the modified section. We also omit the transposition of the bias vectors present in the original algorithm for conciseness.

In these equations, \(m^{l}_{mlp}\) and \(m^{l}_{mha}\) are mask vectors for the \(l\)-th layer of the feed-forward network and the multi-headed attention, respectively. $m^{l}_{mlp} = \mathbf{1}_{D \times 1}$ where $D$ represents the output size of the intermediate layer (the number of rows of $W^{l}_{mlp1}$, or the number of filters as per \citet{pruning}), and $m^{l}_{mha} = \mathbf{1}_{H \times 1}$ where $H$ is the number of attention heads. The concatenated masks can thus be characterized as $m_{mha} =  \mathbf{1}_{H \times L}$ and $m_{mlp} =  \mathbf{1}_{R \times L}$, where $L$ is the total number of layers. As the masks are only used for their representational value, their values stay constant and equal to $1$ throughout the process.

The Fisher information matrix can be formulated in terms of the partial differentials of the loss function \citep{pruning}:
\begin{equation}
\begin{aligned}
I := \frac{1}{|D|} \sum_{(x,y) \in D} \left( \frac{\partial}{\partial m} L(x, y; 1) \right) \left( \frac{\partial}{\partial m} L(x, y; 1) \right)^T
\end{aligned}
\end{equation}

As the full Fisher information matrix takes $O(|\theta|^2)$ to store, it becomes impractical for even a moderate number of masks. Thus, we use the diagonal approximation of the full matrix, as commonly practiced \citep{Kirkpatrick_2017, matena2022merging, pruning}. The masks then have an associated Fisher information value equivalent to the mean squared sum of the gradients:

\begin{equation}
\begin{aligned}
I_{ii} := \frac{1}{|D|} \sum_{(x,y) \in D} \left( \frac{\partial}{\partial m_i} L(x, y; 1) \right)^2
\end{aligned}
\end{equation}

\citet{matena2022merging} provide the following closed-form solution for model merging using the diagonal approximation of the Fisher information $F_j$ for parameters $\theta_j$ of the $j^{th}$ model, with $M$ as the total number of models to be merged:

\begin{equation}
\begin{aligned}
\theta^* = \frac{\sum_{j=1}^{M} \lambda_j F_j  \theta_j} {\sum_{j=1}^{M} \lambda_j F_j}
\end{aligned}
\end{equation}

We keep $\lambda_j = 1$ as we do not conduct hyperparameter tuning on the validation set. We approximate $F_j$ of each parameter with the Fisher information $I_{ii}$ of the mask node $m_i$ according to the following scheme:

\begin{enumerate}
    \item If the parameter is in the query and key matrices of a particular attention head, i.e. $\theta_j$ is in $W_{qk}^{l,h}$ as per Equation \ref{mha}, the Fisher information of the corresponding mask $m_{mha}^{l,h}$ is taken for $F_j$. Including the value parameters here was found to empirically lead to a lower performance.
    \item If the parameter in a particular row $r$ of the intermediate layer of the feed-forward block, or $W^{l, r*}_{mlp1}$ as per Equation \ref{ffd}, the Fisher information of the corresponding mask $m^{l, r}_{mlp}$ is taken taken for $F_j$. 
    \item For all other parameters, $F_j = 1$, resulting in merging equivalent to plain averaging.
\end{enumerate}

We note that our methodology requires the calculation of gradients for only $(H + D) \times L$ parameters, in comparison to Fisher-weighted averaging requiring $|\theta|$, almost certain to be magnitudes higher.

\section{Evaluation}

\begin{table*}
\centering
\begin{tabular}{llllllll}
\hline
\textbf{Model ($\downarrow$)} & \multicolumn{1}{c}{\textbf{Averaging}} & \multicolumn{3}{c}{\textbf{Fisher \citeyearpar{matena2022merging}}} & \multicolumn{3}{c}{\textbf{Ours}} \\
\cline{3-8} 
\textbf{No. of Samples ($\rightarrow$)} & & \textbf{128} & \textbf{2,048} & \textbf{32,768} & \textbf{128} & \textbf{2,048} & \textbf{32,768} \\
\hline
BERT (Base) & 91.2 & 92.0 & 91.7 & 91.9 & \textbf{95.2} {\color{darkgreen}(+4.0)} & 92.8 {\color{darkgreen}(+1.6)} & 93.7 {\color{darkgreen}(+2.5)} \\
\hline
BERT (Large) & 85.8 & 87.0 & 86.1 & 89.2 & \textbf{90.1} {\color{darkgreen}(+4.3)} & 88.1 {\color{darkgreen}(+2.3)} & 87.9 {\color{darkgreen}(+2.1)} \\
\hline
BERT (Tiny) & \textbf{90.8} & 89.6 & 89.2 & 90.4 & 86.1 {\color{darkred}(-4.7)} & 88.4 {\color{darkred}(-2.4)} & 88.7 {\color{darkred}(-2.1)} \\
\hline
RoBERTa & 86.2 & 91.5 & 88.5 & 88.5 & \textbf{92.7} {\color{darkgreen}(+6.5)} & 89.4 {\color{darkgreen}(+3.2)} & 88.8 {\color{darkgreen}(+2.6)} \\
\hline
\end{tabular}
\caption{Normalized and aggregated metrics across tasks.} \label{tab:metrics}
\end{table*}

We evaluate our methodology on multiple pre-trained sizes of BERT, specifically tiny \citep{berttiny}, base, and large \citep{devlin2019bert}, along with the base model for the BERT derivative RoBERTa \citep{liu2019roberta}. We note however that with minimal modification it can be applied to all Transformer-based models.

We use publicly available checkpoints of all models finetuned on particular tasks in the GLUE benchmark \citep{wang-etal-2018-glue}: MNLI \citeplanguageresource{mnli}, SST-2 \citeplanguageresource{sst2}, MRPC \citeplanguageresource{mrpc}, QQP \citeplanguageresource{qqp}, QNLI \citeplanguageresource{squad}, and RTE \citeplanguageresource{rte}. While the datasets we use are wholly in English, no part of our methodology depends on the specific language. However, we assume that for two models being merged, the datasets they were trained on should be in the same language for valid results. We leave investigation of cross-lingual model merging to future work. 

For each checkpoint, we take 128, 2,048, and 32,768 samples from the training set of the corresponding dataset to calculate the Fisher information. We then merge the models, two at a time, and evaluate on the validation set of both the corresponding datasets. As the final classification head layer for each finetuned model is different and task-specific, we use the respective head for each task when evaluating. For comparison, the same procedure is conducted using the original Fisher-weighted merging methodology in \citep{matena2022merging}, along with simple averaging \citep{wortsman2022modelsoup, choshen2022fusing}.

\section{Results} \label{results}

Table \ref{tab:metrics} showcases the performance of our proposed method compared to simple averaging and the Fisher-weighted model merging of \citet{matena2022merging}. For ease of comparison, the difference between our method and the simple averaging is also indicated. Results are normalized based on the respective fine-tuned model's performance on each task, and the median is taken to reduce the effects of outliers. The results demonstrate notable variations based on the model size and the number of samples used for calculating Fisher information.

For BERT (Base), our method demonstrates a consistent improvement over other methods across all sample sizes, with the most significant gain seen at 128 samples, providing a \textbf{+4.0} increase in performance from simple averaging. BERT (Large) and RoBERTa follow a similar trend, with our method consistently outperforming the others. The greatest increase is again at 128 samples, with a performance boost of \textbf{+4.0} and \textbf{+6.5} respectively. 

The results for BERT (Tiny) show an anomalous decrease in performance from simple averaging, in both Fisher-weighted merging and our method. However, we note that our method closes the gap as the number of samples increases, with a -2.1 decrease using 32,768 samples. We also stress the fact that BERT (Tiny) is an exceptionally small model, with only two transformer layers and a hidden embedding size of 128.

We note that our method also provides a significant efficiency advantage over Fisher-weighted averaging, with elapsed time speed-ups of \textbf{66.2x}, \textbf{57.4x}, \textbf{321.7x}, and \textbf{69.5x}, for BERT (Base), BERT (Large), BERT (Tiny), and RoBERTa respectively. To reduce the effect of overhead, the speedup is calculated using only the average runtime of the Fisher information approximation algorithm for each method, with 128 samples, running on a Nvidia T4 GPU.

Overall, these results confirm the efficacy of using the Fisher information of mask nodes for model merging, particularly in scenarios with limited data or computational resources. Further investigation is needed to fully understand the dynamics with scale, particularly for extremely small models like BERT (Tiny).

\section{Conclusion}

In this paper, we introduced a novel model merging method for Transformers that builds upon the Fisher-weighted averaging methodology. By calculating the Fisher information of masks in attention heads and feed-forward layers, our approach significantly reduces the computational cost associated with full-scale Fisher-weighted merging while improving performance. Without a dependence on the validation set, our method is widely applicable for model merging in the context of multi-task learning.

We evaluated our method across several models and tasks, demonstrating its effectiveness and efficiency. The low computation cost enables practical model merging for real-world applications, particularly in resource-constrained environments. 
Future work could be done to evaluate our method's capabilities for merging 3+ models at a time. Novel mask schemes may also be considered, along with an extension to different architectures.

\nocite{*}
\section{Bibliographical References}\label{sec:reference}

\bibliographystyle{lrec-coling2024-natbib}
\bibliography{lrec-coling2024-example}

\begin{thebibliography}{6}
\expandafter\ifx\csname natexlab\endcsname\relax\def\natexlab#1{#1}\fi

\bibitem[{Dolan and Brockett(2005)}]{mrpc}
Dolan, William B and Brockett, Chris. 2005.
\newblock \emph{Microsoft Research Paraphrase Corpus}.
\newblock Microsoft.
\newblock \href {https://www.microsoft.com/en-us/download/details.aspx?id=52398} {[link]}.

\bibitem[{Iyer et~al.(2017)Iyer, Dandekar, , and Csernai}]{qqp}
Shankar Iyer and Nikhil Dandekar and and Kornél Csernai. 2017.
\newblock \emph{Quora Question Pairs}.
\newblock Quora.
\newblock \href {https://quoradata.quora.com/First-Quora-Dataset-Release-Question-Pairs} {[link]}.

\bibitem[{Rajpurkar et~al.(2016)Rajpurkar, Zhang, Lopyrev, and Liang}]{squad}
Rajpurkar, Pranav and Zhang, Jian and Lopyrev, Konstantin and Liang, Percy. 2016.
\newblock \emph{Stanford Question Answering Dataset}.
\newblock Standford Natural Language Processing Group. Association for Computational Linguistics.
\newblock \href {https://rajpurkar.github.io/SQuAD-explorer/} {[link]}.

\bibitem[{Socher et~al.(2013)Socher, Perelygin, Wu, Chuang, Manning, Ng, and Potts}]{sst2}
Socher, Richard and Perelygin, Alex and Wu, Jean and Chuang, Jason and Manning, Christopher D and Ng, Andrew and Potts, Christopher. 2013.
\newblock \emph{Stanford Sentiment Treebank}.
\newblock The Stanford Natural Language Processing Group.
\newblock \href {https://nlp.stanford.edu/sentiment/} {[link]}.

\bibitem[{Wang et~al.(2009)Wang, Singh, Michael, Hill, Levy, and Bowman}]{rte}
Wang, Alex and Singh, Amanpreet and Michael, Julian and Hill, Felix and Levy, Omer and Bowman, Samuel. 2009.
\newblock \emph{Recognizing Textual Entailment Dataset}.
\newblock National Institute of Standards and Technology.
\newblock \href {https://tac.nist.gov/data/RTE/index.html/} {[link]}.

\bibitem[{Williams et~al.(2018)Williams, Nangia, and Bowman}]{mnli}
Williams, Adina and Nangia, Nikita and Bowman, Samuel R. 2018.
\newblock \emph{Multi-Genre Natural Language Inference (MultiNLI) Corpus}.
\newblock Courant Institute of Mathematical Sciences.
\newblock \href {https://cims.nyu.edu/~sbowman/multinli/} {[link]}.

\end{thebibliography}


\begin{thebibliography}{34}
\expandafter\ifx\csname natexlab\endcsname\relax\def\natexlab#1{#1}\fi

\bibitem[{Ainsworth et~al.(2023)Ainsworth, Hayase, and Srinivasa}]{ainsworth2023git}
Samuel~K. Ainsworth, Jonathan Hayase, and Siddhartha Srinivasa. 2023.
\newblock \href {http://arxiv.org/abs/2209.04836} {Git re-basin: Merging models modulo permutation symmetries}.

\bibitem[{Arpit et~al.(2022)Arpit, Wang, Zhou, and Xiong}]{arpit2022ensemble}
Devansh Arpit, Huan Wang, Yingbo Zhou, and Caiming Xiong. 2022.
\newblock \href {http://arxiv.org/abs/2110.10832} {Ensemble of averages: Improving model selection and boosting performance in domain generalization}.

\bibitem[{Cha et~al.(2021)Cha, Chun, Lee, Cho, Park, Lee, and Park}]{cha2021swad}
Junbum Cha, Sanghyuk Chun, Kyungjae Lee, Han-Cheol Cho, Seunghyun Park, Yunsung Lee, and Sungrae Park. 2021.
\newblock \href {http://arxiv.org/abs/2102.08604} {Swad: Domain generalization by seeking flat minima}.

\bibitem[{Choshen et~al.(2022)Choshen, Venezian, Slonim, and Katz}]{choshen2022fusing}
Leshem Choshen, Elad Venezian, Noam Slonim, and Yoav Katz. 2022.
\newblock \href {http://arxiv.org/abs/2204.03044} {Fusing finetuned models for better pretraining}.

\bibitem[{Devlin et~al.(2019)Devlin, Chang, Lee, and Toutanova}]{devlin2019bert}
Jacob Devlin, Ming-Wei Chang, Kenton Lee, and Kristina Toutanova. 2019.
\newblock \href {http://arxiv.org/abs/1810.04805} {Bert: Pre-training of deep bidirectional transformers for language understanding}.

\bibitem[{Entezari et~al.(2022)Entezari, Sedghi, Saukh, and Neyshabur}]{entezari2022role}
Rahim Entezari, Hanie Sedghi, Olga Saukh, and Behnam Neyshabur. 2022.
\newblock \href {http://arxiv.org/abs/2110.06296} {The role of permutation invariance in linear mode connectivity of neural networks}.

\bibitem[{Fifty et~al.(2021)Fifty, Amid, Zhao, Yu, Anil, and Finn}]{fifty2021efficiently}
Chris Fifty, Ehsan Amid, Zhe Zhao, Tianhe Yu, Rohan Anil, and Chelsea Finn. 2021.
\newblock Efficiently identifying task groupings for multi-task learning.
\newblock \emph{Advances in Neural Information Processing Systems}, 34:27503--27516.

\bibitem[{Han et~al.(2021)Han, Zhang, Ding, Gu, Liu, Huo, Qiu, Yao, Zhang, Zhang, Han, Huang, Jin, Lan, Liu, Liu, Lu, Qiu, Song, Tang, Wen, Yuan, Zhao, and Zhu}]{pretrained2}
Xu~Han, Zhengyan Zhang, Ning Ding, Yuxian Gu, Xiao Liu, Yuqi Huo, Jiezhong Qiu, Yuan Yao, Ao~Zhang, Liang Zhang, Wentao Han, Minlie Huang, Qin Jin, Yanyan Lan, Yang Liu, Zhiyuan Liu, Zhiwu Lu, Xipeng Qiu, Ruihua Song, Jie Tang, Ji-Rong Wen, Jinhui Yuan, Wayne~Xin Zhao, and Jun Zhu. 2021.
\newblock \href {https://doi.org/https://doi.org/10.1016/j.aiopen.2021.08.002} {Pre-trained models: Past, present and future}.
\newblock \emph{AI Open}, 2:225--250.

\bibitem[{Hannun et~al.(2021)Hannun, Guo, and van~der Maaten}]{hannun2021measuring}
Awni Hannun, Chuan Guo, and Laurens van~der Maaten. 2021.
\newblock \href {http://arxiv.org/abs/2102.11673} {Measuring data leakage in machine-learning models with fisher information}.

\bibitem[{Ilharco et~al.(2023)Ilharco, Ribeiro, Wortsman, Gururangan, Schmidt, Hajishirzi, and Farhadi}]{taskvectors}
Gabriel Ilharco, Marco~Tulio Ribeiro, Mitchell Wortsman, Suchin Gururangan, Ludwig Schmidt, Hannaneh Hajishirzi, and Ali Farhadi. 2023.
\newblock \href {http://arxiv.org/abs/2212.04089} {Editing models with task arithmetic}.

\bibitem[{Jin et~al.(2023)Jin, Ren, Preotiuc-Pietro, and Cheng}]{jin2023dataless}
Xisen Jin, Xiang Ren, Daniel Preotiuc-Pietro, and Pengxiang Cheng. 2023.
\newblock \href {http://arxiv.org/abs/2212.09849} {Dataless knowledge fusion by merging weights of language models}.

\bibitem[{Kirkpatrick et~al.(2017{\natexlab{a}})Kirkpatrick, Pascanu, Rabinowitz, Veness, Desjardins, Rusu, Milan, Quan, Ramalho, Grabska-Barwinska, Hassabis, Clopath, Kumaran, and Hadsell}]{catastrophic}
James Kirkpatrick, Razvan Pascanu, Neil Rabinowitz, Joel Veness, Guillaume Desjardins, Andrei~A. Rusu, Kieran Milan, John Quan, Tiago Ramalho, Agnieszka Grabska-Barwinska, Demis Hassabis, Claudia Clopath, Dharshan Kumaran, and Raia Hadsell. 2017{\natexlab{a}}.
\newblock \href {https://doi.org/10.1073/pnas.1611835114} {Overcoming catastrophic forgetting in neural networks}.
\newblock \emph{Proceedings of the National Academy of Sciences}, 114(13):3521--3526.

\bibitem[{Kirkpatrick et~al.(2017{\natexlab{b}})Kirkpatrick, Pascanu, Rabinowitz, Veness, Desjardins, Rusu, Milan, Quan, Ramalho, Grabska-Barwinska, Hassabis, Clopath, Kumaran, and Hadsell}]{Kirkpatrick_2017}
James Kirkpatrick, Razvan Pascanu, Neil Rabinowitz, Joel Veness, Guillaume Desjardins, Andrei~A. Rusu, Kieran Milan, John Quan, Tiago Ramalho, Agnieszka Grabska-Barwinska, Demis Hassabis, Claudia Clopath, Dharshan Kumaran, and Raia Hadsell. 2017{\natexlab{b}}.
\newblock \href {https://doi.org/10.1073/pnas.1611835114} {Overcoming catastrophic forgetting in neural networks}.
\newblock \emph{Proceedings of the National Academy of Sciences}, 114(13):3521--3526.

\bibitem[{Kwon et~al.(2022)Kwon, Kim, Mahoney, Hassoun, Keutzer, and Gholami}]{pruning}
Woosuk Kwon, Sehoon Kim, Michael~W. Mahoney, Joseph Hassoun, Kurt Keutzer, and Amir Gholami. 2022.
\newblock \href {https://openreview.net/forum?id=0GRBKLBjJE} {A fast post-training pruning framework for transformers}.
\newblock In \emph{Advances in Neural Information Processing Systems}.

\bibitem[{Li et~al.(2020)Li, Huang, Yang, Wang, and Zhang}]{li2020convergence}
Xiang Li, Kaixuan Huang, Wenhao Yang, Shusen Wang, and Zhihua Zhang. 2020.
\newblock \href {http://arxiv.org/abs/1907.02189} {On the convergence of fedavg on non-iid data}.

\bibitem[{Li et~al.(2016)Li, Yosinski, Clune, Lipson, and Hopcroft}]{li2016convergent}
Yixuan Li, Jason Yosinski, Jeff Clune, Hod Lipson, and John Hopcroft. 2016.
\newblock \href {http://arxiv.org/abs/1511.07543} {Convergent learning: Do different neural networks learn the same representations?}

\bibitem[{Liu et~al.(2021)Liu, Zhang, Kuang, Zhou, Xue, Wang, Chen, Yang, Liao, and Zhang}]{liu2021grouppruning}
Liyang Liu, Shilong Zhang, Zhanghui Kuang, Aojun Zhou, Jing-Hao Xue, Xinjiang Wang, Yimin Chen, Wenming Yang, Qingmin Liao, and Wayne Zhang. 2021.
\newblock Group fisher pruning for practical network compression.
\newblock In \emph{International Conference on Machine Learning}, pages 7021--7032. PMLR.

\bibitem[{Liu et~al.(2019)Liu, Ott, Goyal, Du, Joshi, Chen, Levy, Lewis, Zettlemoyer, and Stoyanov}]{liu2019roberta}
Yinhan Liu, Myle Ott, Naman Goyal, Jingfei Du, Mandar Joshi, Danqi Chen, Omer Levy, Mike Lewis, Luke Zettlemoyer, and Veselin Stoyanov. 2019.
\newblock \href {http://arxiv.org/abs/1907.11692} {Roberta: A robustly optimized bert pretraining approach}.

\bibitem[{Matena and Raffel(2022)}]{matena2022merging}
Michael Matena and Colin Raffel. 2022.
\newblock \href {http://arxiv.org/abs/2111.09832} {Merging models with fisher-weighted averaging}.

\bibitem[{McMahan et~al.(2016)McMahan, Moore, Ramage, Hampson, and y~Arcas}]{McMahan2016CommunicationEfficientLO}
H.~B. McMahan, Eider Moore, Daniel Ramage, Seth Hampson, and Blaise~Ag{\"u}era y~Arcas. 2016.
\newblock \href {https://api.semanticscholar.org/CorpusID:14955348} {Communication-efficient learning of deep networks from decentralized data}.
\newblock In \emph{International Conference on Artificial Intelligence and Statistics}.

\bibitem[{Min et~al.(2021)Min, Ross, Sulem, Veyseh, Nguyen, Sainz, Agirre, Heinz, and Roth}]{min2021recent}
Bonan Min, Hayley Ross, Elior Sulem, Amir Pouran~Ben Veyseh, Thien~Huu Nguyen, Oscar Sainz, Eneko Agirre, Ilana Heinz, and Dan Roth. 2021.
\newblock \href {http://arxiv.org/abs/2111.01243} {Recent advances in natural language processing via large pre-trained language models: A survey}.

\bibitem[{Pascanu and Bengio(2014)}]{pascanu2014revisiting}
Razvan Pascanu and Yoshua Bengio. 2014.
\newblock \href {http://arxiv.org/abs/1301.3584} {Revisiting natural gradient for deep networks}.

\bibitem[{Phuong and Hutter(2022)}]{phuong2022formal}
Mary Phuong and Marcus Hutter. 2022.
\newblock \href {http://arxiv.org/abs/2207.09238} {Formal algorithms for transformers}.

\bibitem[{Qiu et~al.(2020)Qiu, Sun, Xu, Shao, Dai, and Huang}]{pretrained}
XiPeng Qiu, TianXiang Sun, YiGe Xu, YunFan Shao, Ning Dai, and XuanJing Huang. 2020.
\newblock \href {https://doi.org/10.1007/s11431-020-1647-3} {Pre-trained models for natural language processing: A survey}.
\newblock \emph{Science China Technological Sciences}, 63(10):1872--1897.

\bibitem[{Singh and Jaggi(2023)}]{singh2023model}
Sidak~Pal Singh and Martin Jaggi. 2023.
\newblock \href {http://arxiv.org/abs/1910.05653} {Model fusion via optimal transport}.

\bibitem[{Soen and Sun(2021)}]{soen2021variance}
Alexander Soen and Ke~Sun. 2021.
\newblock \href {http://arxiv.org/abs/2107.04205} {On the variance of the fisher information for deep learning}.

\bibitem[{Tatro et~al.(2020)Tatro, Chen, Das, Melnyk, Sattigeri, and Lai}]{tatro2020optimizing}
N.~Joseph Tatro, Pin-Yu Chen, Payel Das, Igor Melnyk, Prasanna Sattigeri, and Rongjie Lai. 2020.
\newblock \href {http://arxiv.org/abs/2009.02439} {Optimizing mode connectivity via neuron alignment}.

\bibitem[{Turc et~al.(2019)Turc, Chang, Lee, and Toutanova}]{berttiny}
Iulia Turc, Ming-Wei Chang, Kenton Lee, and Kristina Toutanova. 2019.
\newblock \href {http://arxiv.org/abs/1908.08962} {Well-read students learn better: On the importance of pre-training compact models}.

\bibitem[{Vaswani et~al.(2017)Vaswani, Shazeer, Parmar, Uszkoreit, Jones, Gomez, Kaiser, and Polosukhin}]{vaswani2017attention}
Ashish Vaswani, Noam Shazeer, Niki Parmar, Jakob Uszkoreit, Llion Jones, Aidan~N Gomez, {\L}ukasz Kaiser, and Illia Polosukhin. 2017.
\newblock Attention is all you need.
\newblock \emph{Advances in neural information processing systems}, 30.

\bibitem[{Wang et~al.(2018)Wang, Singh, Michael, Hill, Levy, and Bowman}]{wang-etal-2018-glue}
Alex Wang, Amanpreet Singh, Julian Michael, Felix Hill, Omer Levy, and Samuel Bowman. 2018.
\newblock \href {https://doi.org/10.18653/v1/W18-5446} {{GLUE}: A multi-task benchmark and analysis platform for natural language understanding}.
\newblock In \emph{Proceedings of the 2018 {EMNLP} Workshop {B}lackbox{NLP}: Analyzing and Interpreting Neural Networks for {NLP}}, pages 353--355, Brussels, Belgium. Association for Computational Linguistics.

\bibitem[{Wortsman et~al.(2022)Wortsman, Ilharco, Gadre, Roelofs, Gontijo-Lopes, Morcos, Namkoong, Farhadi, Carmon, Kornblith, and Schmidt}]{wortsman2022modelsoup}
Mitchell Wortsman, Gabriel Ilharco, Samir~Yitzhak Gadre, Rebecca Roelofs, Raphael Gontijo-Lopes, Ari~S. Morcos, Hongseok Namkoong, Ali Farhadi, Yair Carmon, Simon Kornblith, and Ludwig Schmidt. 2022.
\newblock \href {http://arxiv.org/abs/2203.05482} {Model soups: averaging weights of multiple fine-tuned models improves accuracy without increasing inference time}.

\bibitem[{Yadav et~al.(2023)Yadav, Tam, Choshen, Raffel, and Bansal}]{ties}
Prateek Yadav, Derek Tam, Leshem Choshen, Colin Raffel, and Mohit Bansal. 2023.
\newblock \href {http://arxiv.org/abs/2306.01708} {Resolving interference when merging models}.

\bibitem[{Zhang and Yang(2021)}]{zhang2021survey}
Yu~Zhang and Qiang Yang. 2021.
\newblock A survey on multi-task learning.
\newblock \emph{IEEE Transactions on Knowledge and Data Engineering}, 34(12):5586--5609.

\bibitem[{Zhou et~al.(2023)Zhou, Li, Li, Yu, Liu, Wang, Zhang, Ji, Yan, He, Peng, Li, Wu, Liu, Xie, Xiong, Pei, Yu, and Sun}]{zhou2023comprehensive}
Ce~Zhou, Qian Li, Chen Li, Jun Yu, Yixin Liu, Guangjing Wang, Kai Zhang, Cheng Ji, Qiben Yan, Lifang He, Hao Peng, Jianxin Li, Jia Wu, Ziwei Liu, Pengtao Xie, Caiming Xiong, Jian Pei, Philip~S. Yu, and Lichao Sun. 2023.
\newblock \href {http://arxiv.org/abs/2302.09419} {A comprehensive survey on pretrained foundation models: A history from bert to chatgpt}.

\end{thebibliography}

\section{Language Resource References}
\label{lr:ref}
\bibliographystylelanguageresource{lrec-coling2024-natbib}
\bibliographylanguageresource{languageresource}

\newpage
\part*{Appendix}
\setcounter{section}{0}

\section{Checkpoints}
For all architectures we mention, 6 models finetuned on the 6 respective tasks from the GLUE dataset were used. The checkpoints are given below.
\subsection{BERT Base}
\begin{itemize}
\scriptsize
\item \textbf{MNLI}: \texttt{JeremiahZ/bert-base-uncased-mnli}
\item \textbf{QQP}: \texttt{textattack/bert-base-uncased-QQP}
\item \textbf{QNLI}: \texttt{textattack/bert-base-uncased-QNLI}
\item \textbf{SST2}: \texttt{textattack/bert-base-uncased-SST-2}
\item \textbf{MRPC}: \texttt{textattack/bert-base-uncased-MRPC}
\item \textbf{RTE}: \texttt{textattack/bert-base-uncased-RTE}
\end{itemize}

\subsection{BERT Large}
\begin{itemize}
\scriptsize
\item \textbf{MNLI}: \texttt{yoshitomo-matsubara/bert-large-uncased-mnli}
\item \textbf{QQP}: \texttt{yoshitomo-matsubara/bert-large-uncased-qqp}
\item \textbf{QNLI}: \texttt{yoshitomo-matsubara/bert-large-uncased-qnli}
\item \textbf{SST2}: \texttt{yoshitomo-matsubara/bert-large-uncased-sst2}
\item \textbf{MRPC}: \texttt{yoshitomo-matsubara/bert-large-uncased-mrpc}
\item \textbf{RTE}: \texttt{yoshitomo-matsubara/bert-large-uncased-rte}
\end{itemize}

\subsection{BERT Tiny}
\begin{itemize}
\scriptsize
\item \textbf{MNLI}: \texttt{M-FAC/bert-tiny-finetuned-mnli}
\item \textbf{QQP}: \texttt{M-FAC/bert-tiny-finetuned-qqp}
\item \textbf{QNLI}: \texttt{M-FAC/bert-tiny-finetuned-qnli}
\item \textbf{SST2}: \texttt{M-FAC/bert-tiny-finetuned-sst2}
\item \textbf{MRPC}: \texttt{M-FAC/bert-tiny-finetuned-mrpc}
\item \textbf{RTE}: \texttt{muhtasham/bert-tiny-finetuned-glue-rte}
\end{itemize}

\subsection{RoBERTa}
\begin{itemize}
\scriptsize
\item \textbf{MNLI}: \texttt{JeremiahZ/roberta-base-mnli}
\item \textbf{QQP}: \texttt{JeremiahZ/roberta-base-qqp}
\item \textbf{QNLI}: \texttt{JeremiahZ/roberta-base-qnli}
\item \textbf{SST2}: \texttt{JeremiahZ/roberta-base-sst2}
\item \textbf{MRPC}: \texttt{JeremiahZ/roberta-base-mrpc}
\item \textbf{RTE}: \texttt{JeremiahZ/roberta-base-rte}
\end{itemize}

\section{Compute and Efficiency}
In order to compare the efficiency between our method and full Fisher-weighted merging \citep{matena2022merging}, we calculate the elapsed running times for both, for 128 samples on an Nvidia T4 GPU. As mentioned in Section \ref{results}, only the compute taken for the Fisher information approximation algorithm for both methods is considered: the merging itself is negligible in computation. The times for our method are given in Table \ref{timesours} and for full Fisher calculation in Table \ref{timetheirs}.

We also calculate the FLOPS taken for our method with 128 samples, across all tasks and models, given in Table \ref{gflops}. 

\begin{table}[h]
\centering
\begin{tabular}{ccccc}
\hline
\multicolumn{1}{c}{\textbf{Task ($\downarrow$)}} & \multicolumn{4}{c}{\textbf{Ours, Time (s)}} \\
\hline
\textbf{Model ($\rightarrow$)} & \textbf{Tiny} & \textbf{Base} & \textbf{Large} & \textbf{RoBERTa} \\
\hline
MNLI & 0.056 & 1.121 & 4.591 & 1.195 \\
QQP & 0.063 & 0.670 & 2.405 & 0.630 \\
QNLI & 0.058 & 1.117 & 4.321 & 1.119 \\
SST2 & 0.046 & 0.565 & 3.624 & 0.516 \\
MRPC & 0.048 & 0.972 & 3.624 & 0.970 \\
RTE & 0.053 & 1.251 & 5.047 & 1.307 \\
\hline
\textbf{Mean} & \textbf{0.055} & \textbf{0.959} & \textbf{4.003} & \textbf{0.993} \\
\hline
\end{tabular}
\caption{Time taken for our method with 128 samples across all tasks and models.}
\label{timesours}
\end{table}

\begin{table}[h]
\centering
\begin{tabular}{ccccc}
\hline
\multicolumn{1}{c}{\textbf{Task ($\downarrow$)}} & \multicolumn{4}{c}{\textbf{Fisher \citeyearpar{matena2022merging}, Time (s)}} \\
\hline
\textbf{Model ($\rightarrow$)} & \textbf{Tiny} & \textbf{Base} & \textbf{Large} & \textbf{RoBERTa} \\
\hline
MNLI & 19.149 & 71.486 & 284.276 & 73.203 \\
QQP & 17.472 & 55.193 & 201.505 & 56.623 \\
QNLI & 16.510 & 55.462 & 201.746 & 56.457 \\
SST2 & 16.836 & 56.132 & 203.036 & 57.657 \\
MRPC & 16.376 & 55.973 & 202.999 & 57.304 \\
RTE & 17.011 & 55.851 & 202.527 & 57.856 \\
\hline
\textbf{Mean} & \textbf{17.101} & \textbf{58.009} & \textbf{214.299} & \textbf{59.383} \\
\hline
\end{tabular}
\caption{Time taken for full Fisher calculation with 128 samples across all tasks and models.}
\label{timetheirs}
\end{table}

\begin{table}[h]
\centering
\begin{tabular}{ccccc}
\hline
\multicolumn{1}{c}{\textbf{Task ($\downarrow$)}} & \multicolumn{4}{c}{\textbf{Ours, GFLOPS}} \\
\hline
\textbf{Model ($\rightarrow$)} & \textbf{Tiny} & \textbf{Base} & \textbf{Large} & \textbf{RoBERTa} \\
\hline
MNLI & 21.81 & 4,411.89 & 15,649.25 & 4,366.00 \\
QQP & 12.21 & 2,588.03 & 9,208.30 & 2,521.04 \\
QNLI & 25.08 & 5,008.74 & 17,750.74 & 5,054.74 \\
SST2 & 9.25 & 1,988.07 & 7,080.03 & 1,976.64 \\
MRPC & 19.10 & 3,917.91 & 13,910.27 & 3,963.80 \\
RTE & 29.09 & 5,734.45 & 20,303.87 & 5,734.45 \\
\hline
\textbf{Mean} & \textbf{18.64} & \textbf{4,172.11} & \textbf{14,692.66} & \textbf{4,138.64} \\
\hline
\end{tabular}
\caption{FLOPS for our method with 128 samples across all tasks and models.}
\label{gflops}
\end{table}

\end{document}